\documentclass[conference]{IEEEtran}
\IEEEoverridecommandlockouts
\usepackage{cite}
\usepackage{amsmath,amssymb,amsfonts}
\usepackage{algorithmic}
\usepackage{graphicx}
\usepackage{textcomp}
\usepackage{xcolor}
\usepackage{float}
\def\BibTeX{{\rm B\kern-.05em{\sc i\kern-.025em b}\kern-.08em
    T\kern-.1667em\lower.7ex\hbox{E}\kern-.125emX}}
\usepackage{lipsum}
\usepackage{graphicx}
\ifCLASSOPTIONcompsoc
\usepackage[caption=false, font=normalsize, labelfont=sf, textfont=sf]{subfig}
\else
\usepackage[caption=false, font=footnotesize]{subfig}
\fi

\usepackage{hyperref}
\usepackage{graphicx}
\usepackage{amssymb}
\usepackage{adjustbox}
\usepackage{colortbl}
\usepackage{comment}
\usepackage{multirow}

\definecolor{gray}{cmyk}{0.86,0.86,0.86,0.86}
    
\title{Attention to detail: inter-resolution \\ knowledge distillation}
\begin{document}

\author{
\IEEEauthorblockN{Roc\'io del Amor\IEEEauthorrefmark{1}, Julio Silva-Rodr\'iguez\IEEEauthorrefmark{2}, Adri\'an Colomer\IEEEauthorrefmark{1} \IEEEauthorrefmark{3}, Valery Naranjo\IEEEauthorrefmark{1}}
\IEEEauthorblockA{\IEEEauthorrefmark{1}\textit{Instituto Universitario de Investigaci\'on en Tecnolog\'ia Centrada en el Ser Humano}, \\
\textit{HUMAN-tech, Universitat Polit\`ecnica de Val\`encia}, Valencia, Spain\\
Email: madeam2@upv.es}
\IEEEauthorblockA{\IEEEauthorrefmark{2}\textit{\'ETS Montr\'eal,  Montr\'eal, Qu\'ebec, Canada}}
\IEEEauthorblockA{\IEEEauthorrefmark{3}\textit{ValgrAI- Valencian Graduate School and Research Network for Artificial Intelligence}}
}

\maketitle

\begin{abstract}
The development of computer vision solutions for gigapixel images in digital pathology is hampered by significant computational limitations due to the large  size of whole slide images. In particular, digitizing biopsies at high resolutions is a time-consuming process, which is necessary due to the worsening results from the decrease in image detail. To alleviate this issue, recent literature has proposed using knowledge distillation to enhance the model performance at reduced image resolutions. In particular, soft labels and features extracted at the highest magnification level are distilled into a model that takes lower-magnification images as input. However, this approach fails to transfer knowledge about the most discriminative image regions in the classification process, which may be lost when the resolution is decreased. In this work, we propose to distill this information by incorporating attention maps during training. In particular, our formulation leverages saliency maps of the target class via grad-CAMs, which guides the lower-resolution Student model to match the Teacher distribution by minimizing the \textit{l2}-distance between them. Comprehensive experiments on prostate histology image grading demonstrate that the proposed approach substantially improves the model performance across different image resolutions compared to previous literature. The project code is available on \url{https://github.com/cvblab/kd_resolution}.
\end{abstract}


\section{Introduction}
\label{sec:intro}

Computer vision methods using deep learning have reached remarkable  results in a wide range of applications, including medical image analysis. This is the case even in challenging fields such as digital pathology,  where digitized biopsies take the form of whole-slide images (WSI) consisting of hundreds of thousands of pixels. In addition, the relevant areas may be contained in a small region of the image. Different successful applications of deep learning in digital pathology include global tasks such as biopsy level cancer detection \cite{Campanella2019}, tissue segmentation \cite{weglenet} or small structures detection, such as mitotic figures \cite{Ciresan2013}. In real-world scenarios, the highest image augmentations might not be available, and hardware and time constraints can present challenges to deploying these models. These limitations emphasize the need for novel and efficient solutions for implementing computer-aided systems into clinical practice. In this line, recent work has explored the feasibility of using well-known knowledge distillation formulations to reduce the required image resolution during inference \cite{DiPalma2021}. However, the vanilla knowledge distillation and feature-matching terms used in these studies do not allow for transferring relevant and discriminative regions utilized in high-resolution images. Based on these observations, we propose an attention-aware formulation to reduce the required image resolution during model deployment. The key contributions of our work can be summarized as follows:

\begin{itemize}
    \item A novel attention-constrained formulation for inter-resolution knowledge distillation.
    
    \item We propose to train a Student model which matches the attention maps produced by a Teacher model trained with higher-resolution images by minimizing the \textit{l2}-distance.

    \item In particular, we propose to transfer  only \textit{strictly-positive} gradients in the proposed AT$^{+}$ term. 

    
    \item The method is validated in the context of prostate histology image grading. Using the proposed term, the model archives competitive results while requiring $8\times$ fewer augmentations during deployment.

\end{itemize}

\section{Related Work}
\label{sec:rw}

\paragraph*{\textbf{Constrained classification}} Constrained learning aims to regularize the training of deep learning models for image classification tasks to produce a solution that satisfies a given condition. Through this condition, the model can incorporate additional prior knowledge of the task. The main core of literature in this field is the introduction of constraints at the pixel-level response of the model, which is achieved through the regularization of attention maps. It is worth mentioning that, despite the recent widespread use of attention for the trainable blocks of popular Transformers encoders \cite{NIPS2017_3f5ee243}, we refer to attention as the saliency class-activation maps produced by the image-level classifier when applied at the pixel-level, in the form of semantic segmentation maps or processed response maps such as Grad-CAMs \cite{Selvaraju2017}. In this fashion, weakly supervised segmentation models can introduce object proportion constraints as priors \cite{Pathak2015, proportionConstrainedJS, rocioColitis}, or unsupervised anomaly detection models can be forced to focus on the whole image \cite{bmvc_js, media_js_anomaly, Venkataramanan2020}. Also, self-supervised training methods can leverage the most discriminant regions to enhance feature learning \cite{Selvaraju2021}, or attention constraints can be used to improve fine-grained image recognition \cite{Sun2018}. 

\paragraph*{\textbf{Knowledge distillation}} Knowledge distillation \cite{Hinton2015} is a field of machine learning that explores the use of deep learning solutions in creating efficient models. The most popular scenario aims to reduce the model size for deployment in systems with limited resources, so-called model compression. The main core of the literature involves using a pre-trained Teacher model, trained on large-capacity settings, to transfer an optimal solution to a low-capacity Student model. Vanilla knowledge distillation (KD) \cite{Hinton2015} transfers the softmax scores of the Teacher to the Student, which enhances inter-class relations. Other popular terms match intermediate feature representations (FM) \cite{Romero2015} between both models via an \textit{l2}-loss. Recent works \cite{Yang2021} have improved the base feature representation matching by indirectly using the Teacher classifier through Student features by a softmax regression (SR) of the produced logits. However, these terms do not ensure that the Student model focuses on the same image regions for classification. To alleviate this issue, \cite{Zagoruyko2017} introduced an attention transfer loss (AT), which forces the Student model to match the attention maps produced by the Teacher. Concretely, the gradients obtained for each spatial feature representation (before flattening or global pooling) concerning the logits output by the model are used as a proxy for attention distillation. Nevertheless, as later pointed out in \cite{Selvaraju2017}, gradients alone might produce suboptimal results for highlighting the most relevant patterns for the task at hand. In this work, we refine the AT term to (i) weight both feature representation and gradients, and (ii) use only \textit{strictly positive} gradients, based on Grad-CAMs for attention generation \cite{Selvaraju2017}. 

Besides model compression, knowledge distillation has also been successfully applied to other computer vision fields such as semi-supervised learning \cite{Xie2020} or multi-modal to mono-modal segmentation \cite{Hu2020}. This work focuses on applying knowledge distillation to enable the efficient use of deep learning models at lower image resolutions, which we refer to as inter-resolution knowledge distillation  \cite{DiPalma2021}. Despite its importance in high-demanding image resolution fields such as digital pathology, there is limited literature in this area, and only basic vanilla KD or feature-matching distillation has been validated \cite{DiPalma2021}.

\section{Methods}
\label{sec:methods}

\begin{figure*}[h!]
\begin{center}
\includegraphics[width=0.75\textwidth]{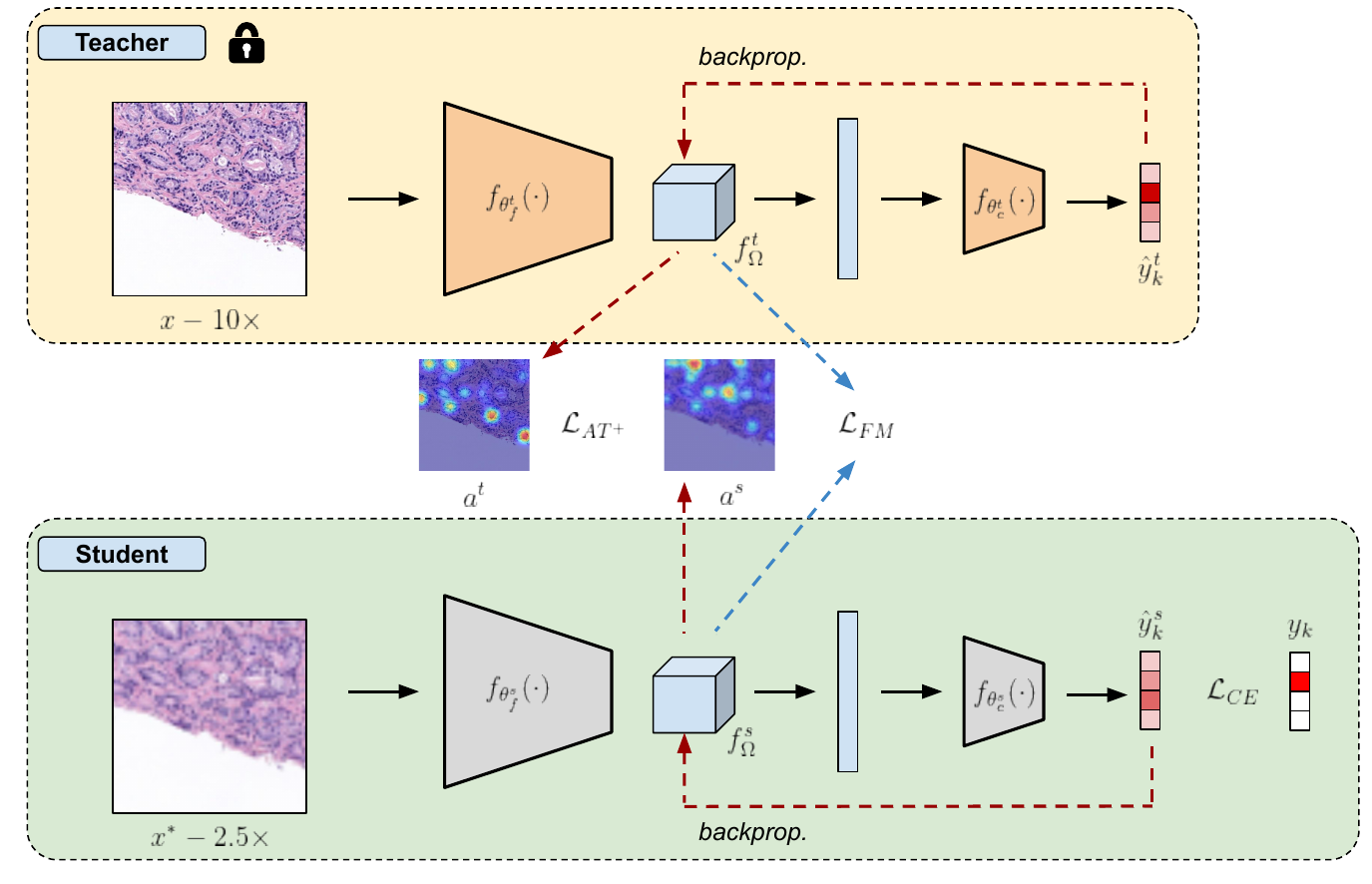}
\vspace{-1 em}
\caption{\textbf{Method overview}. In the context of inter-resolution knowledge distillation, a Teacher model is trained using high-resolution images by optimizing Eq.\ref{eq:ce}. To enable the deployment of efficient models that can operate at low resolutions, we train a Student model by transferring the information from the frozen Teacher. We use the well-known feature-matching distillation (Eq.\ref{eq:fm}) and propose a novel attention-matching term, $AT^{+}$ (Eq.\ref{eq:fm}), which distills spatial information of relevant regions in the image by using strictly positive gradient weighting for attention generation (Eq.\ref{eq:am}). Both terms are combined with the standard cross-entropy loss (Eq. \ref{eq:all}) for the optimization of the Student model.}
\label{fig:summary}
\end{center}
\vspace{-4mm}
\end{figure*}

An overview of our proposed method is depicted in Figure \ref{fig:summary}. In the following, we describe the problem formulation and each of the proposed components.

\paragraph*{\textbf{Preliminaries}} In the context of image classification, we denote a neural networks classifier as $\theta=\{\theta_{f}, \theta_{c}\}$. The model is composed of a convolutional feature extractor $\theta_{f}$, which processes input images $x$ to extract a compressed pixel-level feature representation, $f_{\Omega} \in \ \mathbb{R}^{C \times \Omega}$, such that $C$ represents the manifold dimensionality, and $\Omega$ is the spatial size. The classifier $\theta_{c}$ then utilizes the global-average pooling operation on the feature representation to output softmax scores $\hat{y}_k$ for the $K$ target categories. In a standard image classification scenario, the model is optimized to minimize the cross-entropy loss between predicted scores and labels $y_k$ as follows:

\begin{equation}
\label{eq:ce}
\mathcal L_{CE} = -\frac{1}{K}\sum_{k=1}^{K} y_{k}log(\hat{y}_k)
\end{equation}

\subsection{Inter-resolution knowledge distillation}

Let us denote a Teacher model as $\theta^{t}$, which is trained using standard cross-entropy, as shown in Eq. \ref{eq:ce}, on high-resolution input images ($x$). The objective of inter-resolution knowledge distillation is to train a Student model $\theta^{s}$, which takes distorted input images at a lower resolution ($x^*$) to take into account the rich information contained in the frozen Teacher model to improve its performance. Vanilla knowledge distillation \cite{Hinton2015} distills softmax outputs from Teacher model to incorporate image-specific inter-class dependencies by cross entropy such as:

\begin{equation}
\label{eq:kd}
\mathcal L_{KD} = -\frac{1}{K}\sum_{k=1}^{K} \hat{y}^t_k log( \hat{y}^s_k)
\end{equation}

Since this term only provides global information, the feature matching term in \cite{Romero2015} allows the distilling of information regarding the feature representation by minimizing the l2-distance of both embeddings. Applied before the global average pooling, this matching is applied spatially such that: 

\begin{equation}
\label{eq:fm}
\mathcal L_{FM} = \frac{1}{|\Omega|}\sum_{i\in \Omega}||f_{i}^t - f_{i}^s||^2
\end{equation}

\subsection{Attention matching} 

The  $\mathcal  L_{FM}$ term does not leverage the most relevant regions for the target class classification. This limitation was addressed in \cite{Zagoruyko2017} by matching attention maps obtained solely by the gradients of each spatial feature representation concerning the target class. Nevertheless, later studies on attention generation \cite{Selvaraju2017} suggested using only strictly positive gradients, weighted according to the feature space, to obtain refined attention maps. Thus, we compute attention maps using grad-CAMs \cite{Selvaraju2017}, defined as $a_{\Omega,k}= ReLU(\sum_{c} \alpha_{c,k} f_{\Omega,c})$, where $\alpha_{c}$ represents the class-wise generated gradients and are defined as $\alpha_{c,k}=\frac{\partial f_{c}}{\partial \hat{y}^s_k}$, using the logits before the softmax activation as the target. Then, the attention information is distilled into the Student training using the proposed AT$^{+}$ term as follows:

\begin{equation}
\label{eq:am}
\mathcal L_{AT^{+}} = \frac{1}{K}\sum_{k}\frac{1}{|\Omega|}\sum_{i\in \Omega}||a_{i,k}^t - a_{i,k}^s||^2
\end{equation}

Based on the empirical observations, which are detailed in the experimental section, we include in our formulation the $\mathcal  L_{FM}$ and $\mathcal L_{AT^{+}}$ terms, and we propose to train the Student using the following global criteria:

\begin{equation}
\label{eq:all}
\mathcal L = \mathcal L_{CE} + \alpha_{FM} \mathcal L_{FM}  + \alpha_{AT^{+}} \mathcal L_{AT^{+}}
\end{equation}

\noindent $\alpha \in \mathbb{R}^+$ is the relative weight of each knowledge distillation term, which weights its importance with respect the standard cross entropy loss in Eq. \ref{eq:ce}.

\section{Experiments}
\label{sec:experiments}

\subsection{Experimental setting}

\paragraph*{\textbf{Datasets}} The methods described in this work are validated in patch-level cancerous histology image grading. In particular, we use SICAPv2 \cite{SICAPv2}, a public prostate histology image dataset for patch-level Gleason grading. The dataset includes $10,340$ tissue patches of size $512$ pixels at $10\times$ magnification. According to the presence of cancerous tissue and its severity, images have been labeled by expert pathologists using the following labels: non-cancerous tissue, Gleason grade $3$, $4$, and $5$. The dataset presents class-balanced training, validation, and testing splits, divided at the patient level. 

\paragraph*{\textbf{Implementation Details}} First, Teacher model is trained using images at $10\times$ magnification via standard cross-entropy loss defined in Eq. \ref{eq:ce}. Then, the Student model is trained for different lower-magnifications inputs by optimizing the knowledge distillation criteria in Eq. \ref{eq:kd}, using the Student with frozen weights. We use images at $5\times$, $2.5\times$, and $1.25\times$ magnification as input. Low-magnification inputs are artificially created from the original $10\times$ magnification images by sequential resampling and bilinear interpolating. Teacher and Student models are initialized with the same feature extractor, VGG16 pre-trained on Imagenet. The models are trained during $20$ epochs using ADAM optimizer, with a learning rate of $1e-4$ and a batch size of $32$ images. Data augmentation is incorporated in random image rotations, color jitter, and random affine transforms. During training, we balance the class distribution of the training samples using proportional sampling. We monitor the performance of the model on the validation set throughout training and select the best-performing model for testing. To determine the relative weights of each knowledge distillation term, we follow the same validation procedure with different values of $\alpha=\{0, 0.01, 0.1, 1, 10, 100\}$.
\paragraph*{\textbf{Baselines}} To validate the goodness of the proposed method, we focus on relevant previous methods of knowledge distillation. Concretely, we use the vanilla knowledge distillation over softmax outputs (KD) \cite{Hinton2015}, and its optimization in combination with the feature matching term (KD + FM) \cite{DiPalma2021}. Also, we include softmax regression (FM + SR) by matching the logits produced by the features of Teacher and Student models through the frozen Teacher, as described in \cite{Yang2021}. It is worth mentioning that only KD \cite{Hinton2015} and KD + FM \cite{DiPalma2021} have been previously proposed in the context of inter-resolution knowledge distillation. Still, we include FM + SR \cite{Yang2021}, which obtains leading results for model compression, the core of the knowledge distillation literature. Also, we use the Student models trained at different resolutions without any knowledge distillation as a baseline. All models are trained using the same hyperparameter setting described above for the proposed method, which showed consistent performance.

\paragraph*{\textbf{Evaluation Metrics}} We use standard metrics used in previous literature for disease grading in the medical context. In particular, we use the accuracy (Acc) between predicted and reference labels and Cohen's quadratic kappa ($\kappa$), which considers inter-rater agreement in the context of ordered categories classification. To ensure reproducibility and account for random events in weight initialization, we repeat all experiments using three fixed seeds and average the results across repetitions.

\subsection{Results}
\label{sec:results}


\paragraph*{\textbf{Comparison to the literature}}

The results obtained using the proposed method and baseline approaches on the test subset are presented in Table \ref{table:soa comparison}. Although using the vanilla KD \cite{Hinton2015}  yields the best results when using  $5\times$ augmentation images, its performance deteriorates as the resolution decreases and the information becomes more distorted. In addition, feature regularization via FM \cite{DiPalma2021} or SR \cite{Yang2021} terms does not show relevant improvements over KD in inter-resolution knowledge distillation. In contrast, the proposed attention matching (AT$^{+}$) methodology performs comparably to these methods at $5\times$ magnification and outperforms previous literature by $\sim 3\%$ accuracy for images at $2.5\times$, and $1.25\times$ magnification. In addition, the proposed AT$^{+}$ formulation can obtain comparable accuracy to the Teacher model, trained at $10\times$ magnification, with even $8$ times less resolution.

\begin{table}[h!]
\footnotesize
\centering
\caption{Comparison to previous literature on SICAPv2. The metric presented is the accuracy and quadratic kappa (Acc/$\kappa$).}
\label{table:soa comparison}
\resizebox{1\linewidth}{!}{
\begin{tabular}{|l|cccc|}
\hline
\multicolumn{1}{|c|}{\textbf{Method}} &
  \multicolumn{4}{c|}{\textbf{Augmentation}} \\ \hline
\multicolumn{1}{|c|}{\textbf{}} &
  \multicolumn{1}{c|}{\textbf{10x}} &
  \multicolumn{1}{c|}{\textbf{5x}} &
  \multicolumn{1}{c|}{\textbf{2.5x}} &
  \textbf{1.25x} \\ \hline
Teacher &
  \multicolumn{1}{c|}{0.733/0.829} &
  \multicolumn{1}{c|}{0.723/0.781} &
  \multicolumn{1}{c|}{0.710/0.789} &
  0.700/0.772 \\ \hline
KD \cite{Hinton2015} &
  \multicolumn{1}{c|}{-} &
      \multicolumn{1}{c|}{\textbf{0.778/0.845}} &
  \multicolumn{1}{c|}{0.743/0.803} &
  \multicolumn{1}{c|}{0.700/0.772} \\ \hline
FM + KD \cite{DiPalma2021}&
  \multicolumn{1}{c|}{-} &
  \multicolumn{1}{c|}{0.741/0.828} &
  \multicolumn{1}{c|}{0.733/0.806} &
  0.709/0.787 \\ \hline
FM + SR \cite{Yang2021} &
  \multicolumn{1}{c|}{-} &
  \multicolumn{1}{c|}{0.754/0.841} &
  \multicolumn{1}{c|}{0.742/0.799} &
  \multicolumn{1}{c|}{0.694/0.725} \\ \hline
  \rowcolor{gray!4}
FM + AT$^{+}$ &
  \multicolumn{1}{c|}{-} &
  \multicolumn{1}{c|}{0.763/0.837} &
  \multicolumn{1}{c|}{\textbf{0.770/0.807}} &
  \textbf{0.731/0.803} \\ \hline
\end{tabular}
}
\end{table}

In the following, we provide comprehensive ablation experiments to validate the different elements of the proposed methodology. It is worth mentioning that the results further presented for the ablation experiments are obtained on the validation set.

\begin{figure}[h!]
    \begin{center}
    \includegraphics[width=1\linewidth]{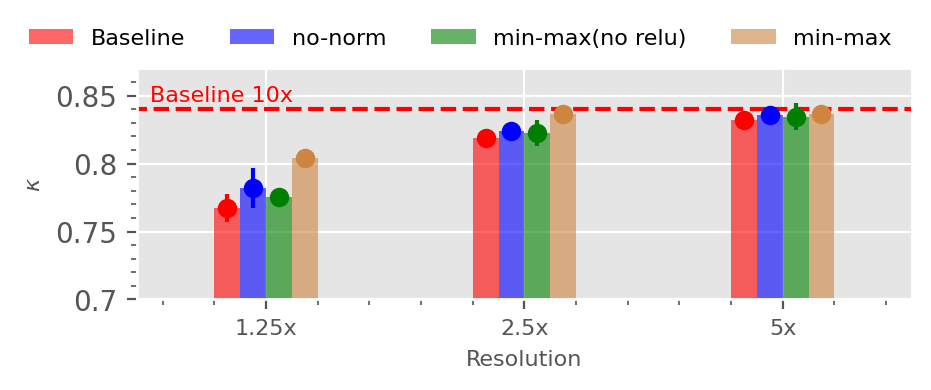}
    \end{center}
    \vspace{-2 em}
    \caption{Ablation study of the effect of attention map normalization on the method performance. The Teacher model trained at the different resolutions is used as a baseline.}
    \vspace{-1 em}
     \label{fig:ablation_normalization}
\end{figure}

\paragraph*{\textbf{Optimizing AT$^{+}$ distillation}} First, we study the best way of leveraging normalized attention maps for knowledge distillation. To this end, after the gradient-weight of feature maps in Eq. \ref{eq:am} to obtain pixel-level logits, we study different settings: no normalization, min-max normalization to clip logits to $[0, 1]$ values, and min-max normalization after ReLU activation, which weights only regions with positive gradients for each target class, as used in Grad-CAM \cite{Selvaraju2017}. Results are depicted in Figure \ref{fig:ablation_normalization}. The results in the validation show that using ReLU activation is essential for the correct knowledge distillation at lower resolutions, with improvements over other attention computations such as the one used in \cite{Zagoruyko2017}. Also, normalizing the obtained logits improves the model optimization. This can be explained due to it allows both the Teacher and the Student to reach absolute values for the embedding space without affecting the knowledge transfer. It is worth mentioning that this term is only applied during training, so the proposed formulation does not involve any additional computational burden during inference.

\paragraph*{\textbf{Combination with other knowledge distillation terms}} Next, we depict studies to evaluate the combination of the proposed attention matching (AT$^{+}$) loss with prior popular terms on knowledge distillation. Concretely, the results obtained in the validation subset of combining AT$^{+}$ with vanilla knowledge distillation (KD) and feature matching (FM) are presented in Table \ref{table:ablation_kd_terms}. Results show how the FM term is the only one that produces improvements over the AT$^{+}$ criteria alone, which is accentuated at lower augmentations.

\begin{table}[h!]
\footnotesize
\centering
\caption{Ablation experiment on the effect of the combination of different knowledge distillation terms. The metric presented is the accuracy and quadratic kappa (Acc/$\kappa$).}
\label{table:ablation_kd_terms}
\begin{tabular}{|l|ccc|}
\hline
\multicolumn{1}{|c|}{\textbf{Method}} & \multicolumn{3}{c|}{\textbf{Augmentation}}                                                                                  \\ \hline
\multicolumn{1}{|c|}{\textbf{}}       & \multicolumn{1}{c|}{\textbf{5x}}               & \multicolumn{1}{c|}{\textbf{2.5x}}             & \textbf{1.25x}            \\ \hline
Teacher                              & \multicolumn{1}{c|}{0.746/0.831} & \multicolumn{1}{c|}{0.740/0.818} & 0.706/0.767 \\ \hline
AT$^{+}$                                    & \multicolumn{1}{c|}{0.773/0.836} & \multicolumn{1}{c|}{0.765/0.836} & 0.764/0.804 \\ \hline
AT$^{+}$ + KD                               & \multicolumn{1}{c|}{0.753/0.846} & \multicolumn{1}{c|}{0.761/0.834} & 0.729/0.781 \\ \hline
AT$^{+}$ + FM & \multicolumn{1}{c|}{\textbf{0.767/0.849}} & \multicolumn{1}{c|}{\textbf{0.776/0.837}} & \textbf{0.753/0.812} \\ \hline
AT$^{+}$ + KD + FM                          & \multicolumn{1}{c|}{0.761/0.840} & \multicolumn{1}{c|}{0.775/0.836} & 0.749/0.789 \\ \hline
\end{tabular}
\end{table}

\begin{figure}[h!]
    \begin{center}
    \includegraphics[width=1.\linewidth]{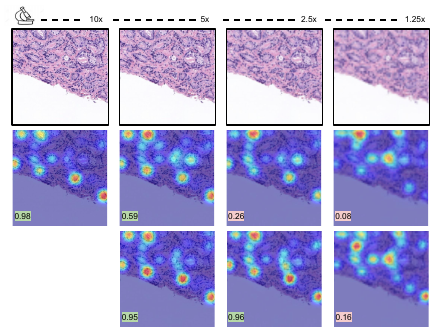}
    \end{center}
    \vspace{-1 em}
    \caption{Qualitative assessment of the effect of the attention matching (AT$^{+}$) term. The top row presents original images at different resolution levels (augmentations). The second row shows the Student output to the target class trained without any knowledge distillation and the attention map produced. The last row shows the effect of distilling the knowledge from the Teacher model trained at 10x magnification. Green probabilities indicate a correctly classified sample, while red indicates the opposite. }
    \label{fig:qualitative}
\end{figure}

\paragraph*{\textbf{Qualitative evaluation}} Finally, we include qualitative visualization of the effect of including the proposed attention matching in the training of the low-resolution Student, see Figure \ref{fig:qualitative}. In the context of tumor grading, for a test case with Gleason grade 3, the Teacher model's attention (shown in the second row on the left) focuses on small individual glands. However, as the input image quality deteriorates, the model shifts its focus to different patterns, resulting in incorrect classifications. Upon incorporating the AT$^{+}$ term in the Student training (shown in the third row), the model's attention aligns with that of the Teacher model, focusing on the same glandular patterns while simultaneously improving the output score for the correct class.

\vspace{-1mm}
\section{Conclusions}
\label{sec:conclusions}

This work introduces a novel constrained formulation for knowledge distillation, aimed at enabling efficient image classification at lower resolutions. Specifically, we propose to distill the knowledge from a high-resolution Teacher model on the localization of discriminative regions in images for the given task. To achieve this, we formulate the attention matching loss, AT$^{+}$, which forces the Student model trained at low resolution to focus on the same regions as the Teacher model by minimizing the \textit{l2}-distance between attention maps obtained using Grad-CAMs. The proposed approach is successfully validated in the context of histology image grading. In this field, the large size of digitized biopsies and the large augmentation required are a burden for applying deep learning solutions on real-time computer-aided diagnostic systems. The obtained results show that attention distillation allows operating at up to $8\times$ fewer augmentations and outperforms previous relevant literature in knowledge distillation in $\sim 3\%$ accuracy. Still, it is worth mentioning that the performance of these methods might be limited at some minimum resolution since any of the baselines and proposed methods performed properly at resolutions below $0.625\times$ augmentations. We believe that the results obtained open further research directions to allow the efficient use of deep learning solutions in the medical context.

\section*{Acknowledgement}
This work has received funding from Horizon 2020, the European Union's Framework Programme for Research and Innovation, under the grant agreement No. 860627 (CLARIFY), the Spanish Ministry of Economy and Competitiveness through project PID2019-105142RB-C21 (AI4SKIN) and  GVA through the project INNEST/ 2021/321 (SAMUEL). The work of Roc\'io del Amor has been supported by the Spanish Ministry of Universities (FPU20/05263).  The work of J. Silva-Rodr\'iguez was carried out during his previous position at Universitat Polit\`ecnica de Val\`encia. The work of Adri\'an Colomer has been supported by the ValgrAI – Valencian Graduate School and Research Network for Artificial Intelligence and Generalitat Valenciana and Universitat
Politecnica de Valencia (PAID-PD-22).

\bibliography{refs.bib}
\bibliographystyle{IEEEtran}

\end{document}